\documentclass[openacc]{rstransa}

\usepackage{bm}
\usepackage{mathrsfs}

\newcommand{\dl}{{\mathscr{L}}}
\newcommand{\xs}{{\mathbf{x}}}

\newcommand{\thetas}{{\bm{\theta}}}

\titlehead{Research}

\begin{document}

\title{Bayesian symbolic regression: Automated equation discovery from a physicists' perspective}

\author{
Roger Guimerà$^{1, 2}$ and Marta Sales-Pardo$^{1}$
}

\address{
$^{1}$Department of Chemical Engineering, Universitat Rovira i Virgili, 43007 Tarragona, Catalonia\\
$^{2}$ICREA, 08010 Barcelona, Catalonia
}

\subject{xxxxx, xxxxx, xxxx}


\corres{Roger Guimerà; Marta Sales-Pardo\\
\email{roger.guimera@urv.cat; marta.sales@urv.cat}}

\begin{abstract}
Symbolic regression automates the process of learning closed-form mathematical models from data. Standard approaches to symbolic regression, as well as newer deep learning approaches, rely on heuristic model selection criteria, heuristic regularization, and heuristic exploration of model space. Here, we discuss the probabilistic approach to symbolic regression, an alternative to such heuristic approaches with direct connections to information theory and statistical physics. We show how the probabilistic approach establishes model plausibility from basic considerations and explicit approximations, and how it provides guarantees of performance that heuristic approaches lack. We also discuss how the probabilistic approach compels us to consider model ensembles, as opposed to single models.
\end{abstract}


\begin{fmtext}

\section{Introduction}

It took four years for Kepler to establish that Mars' orbit is elliptical, in 1609; and it was not until 1687 that Newton unified his empirical observations into a mathematical model. Can we design computer programs and theoretical frameworks to automate this process and make it faster? Can machine learning revolutionize science as it is revolutionizing other areas of our lives? \cite{wang2023,cornelio23} At least since the 1970s, some researchers have thought so, and have tried to develop algorithms that automatically learn closed-form models from data \cite{langley87,dzeroski07}. Under diverse names such as computational scientific discovery, equation discovery or, more recently, symbolic regression, this field has grown, matured and is becoming established within machine learning.

\end{fmtext}
\maketitle

Traditional symbolic regression uses genetic algorithms to evolve populations of expressions that are `well adapted` to the data at hand \cite{koza92,schmidt09,pysr}, that is, expressions that: (i) describe the data well; and (ii) are reasonably simple. To operationalize these two criteria, two loss functions are heuristically defined, one for error and one for complexity. They are eventually combined, again heuristically, leading to a single unified model-selection criterion. More modern approaches have recently been proposed based on sparse regression \cite{brunton16}, recurrent neural networks \cite{petersen2021}, variational autoencoders \cite{meznar2023}, or a combination of neural networks  with physics-inspired techniques \cite{udrescu20}, among others (for systematic reviews, see Refs.~\cite{cava2021,makke2024}). 
However, despite their differences, these modern methods share with traditional symbolic regression (at least to some degree) the need to define loss functions and model selection criteria heuristically. The way they explore model space and come up with specific models is also heuristic.

Here, we discuss an alternative approach. Rather than comparing methods based on their performance on benchmark problems or datasets (which are inevitably limited and biased, and eventually lead to methods overfitting the benchmarks), we argue that symbolic regression approaches should conform to some basic desiderata. In particular, we demand from symbolic regression approaches the following properties:
\begin{enumerate}
    \item They must establish the plausibility of models based on rigorous arguments and, when necessary, explicit---and hence scrutinizable---assumptions and/or approximations.
    \item They must integrate goodness of fit and model complexity into a single measure of plausibility, so that no {\em ad hoc} parameters or thresholds need to be fixed to balance them.
    \item They must be consistent, that is, they must select the true model with probability approaching one as the sample size grows to infinity.
    \item They must account for the uncertainty inherent in the model discovery process.
\end{enumerate}
We show that a probabilistic (or Bayesian) approach to symbolic regression \cite{guimera20,reichardt20,fajardo23} satisfies all of this conditions. This approach draws from probability theory, information theory and statistical physics and, we believe, provides a solid foundation for  future developments in the area.

\section{Bayesian symbolic regression}

In symbolic regression, we aim to identify the closed-form mathematical model $m^*(\xs, \thetas^*)$ that is responsible for the generation of some observed dependent variables $\{y^k\}$ through the process
\begin{align}
y^k=m^*(\xs^k, \thetas^*) + \epsilon^k \;.
\label{eq:datagen}
\end{align}
Here, $\xs^k$ is the $k$-th observation of the features or independent variables, $\thetas^*$ are some parameters of the model $m^*$, and $\epsilon^k$ is an observational noise, assumed to be Gaussian-distributed with zero mean and unknown variance $\sigma^2$. The space of candidate models $\mathcal{M}$ comprises, in principle, all possible closed-form models $m_i(\xs, \thetas_i)$, although in some situations one may want to restrict the space to certain subsets of models.

Given that there is uncertainty in both the data generation process  and in the model selection itself, the most complete description of the symbolic regression problem is probabilistic. Indeed, given some observed data $D=\{(y^k, \xs^k), k=1, \dots, N\}$, the complete solution to the problem is given by the conditional probability  $p(m_i|D)$ of $m_i$ being the true generating model given the data $D$. Indeed, given this distribution $p(m_i|D)$ over models $m_i \in \mathcal{M}$, we can answer any model-selection question (for example, what is the most plausible model?) or make any prediction (for example, what is the probability that $y$ is larger than a certain value at some point $\xs$?). The practical question is, then, whether $p(m_i|D)$ can be computed or, at least, approximated.

The answer to this question is that, under relatively mild approximations, $p(m_i|D)$ can indeed be computed. First, consider the joint distribution $p(m_i, \thetas_i|D)$ of the model $m_i$ and its parameters $\thetas_i$, given $D$. This distribution can be written in terms of the likelihood $p(D|m_i, \thetas_i)$ by application of Bayes' theorem
\begin{align}
p(m_i, \thetas_i|D) = \frac{p(D|m_i, \thetas_i) \, p(m_i, \thetas_i)}{p(D)} = \frac{p(D|m_i, \thetas_i) \, p(\thetas_i|m_i) \, p(m_i)}{p(D)} \;.
\end{align}

In turn, since by hypothesis data are generated according to Eq.~\eqref{eq:datagen}, the likelihood is
\begin{eqnarray}
p(D|m_i, \thetas_i) & = & \prod_{k=1}^{N}\frac{1}{\sqrt{2\pi \sigma^2}} \exp \left[ -\frac{\left(y^k - m_i(\xs^k, \thetas_i) \right)^2}{2\sigma^2}\right] \nonumber \\
& = & \frac{1}{\left(2\pi \sigma^2\right)^{N/2}} \exp \left[ -\frac{\sum_{k=1}^{N}\left(y^k - m_i(\xs^k, \thetas_i) \right)^2}{2\sigma^2}\right] \; .
\end{eqnarray}
Putting it all together, one can calculate the posterior distribution $p(m_i|D)$ by marginalizing over parameter values\footnote{Note that the unknown variance $\sigma^2$ of the observational noise is also a parameter of the probabilistic model, although it is not a parameter of $m_i$ itself. In a slight abuse of notation, in the following integrals we include $\sigma$ into $\thetas_i$ so as to keep expressions a bit more concise.} 
\begin{eqnarray}
p(m_i | D) & = & \int_{\Theta_i} {\rm d}\thetas_i \, p(m_i, \thetas_i | D) \nonumber \\
& = & \frac{p(m_i)}{p(D)} \int_{\Theta_i} {\rm d}\thetas_i \, p(D|m_i, \thetas_i) \, p(\thetas_i|m_i) \nonumber \\
& = & \frac{p(m_i)}{p(D) \left(2\pi \sigma^2\right)^{N/2}} \int_{\Theta_i} {\rm d}\thetas_i \, \exp \left[ -\frac{\sum_{k=1}^{N}\left(y^k - m_i(\xs^k, \thetas_i) \right)^2}{2\sigma^2}\right] \, p(\thetas_i|m_i) \nonumber \\
& = & \frac{\exp [-\dl(m_i,D)]}{Z}\; ,
\label{eq:posterior}
\end{eqnarray}
where the last step is simply notation and can be regarded as the definition of $\dl(m_i, D)$, and $Z=p(D)$ is introduced to make it clear that, in the context of model selection, $p(D)$ is just a normalizing constant.

In general, the integral in Eq.~\eqref{eq:posterior} cannot be evaluated analytically because the model $m_i(\xs, \thetas_i)$ and the prior $p(\thetas_i|m_i)$ may have arbitrarily complex dependencies on the parameters $\thetas_i$. However, the integral can be estimated by means of the Laplace approximation by assuming that: (i) the likelihood is sufficiently peaked around the parameter values $\hat{\thetas}_i$ that maximize the likelihood; (ii) the prior $p(\thetas_i|m_i)$ is sufficiently smooth around $\hat{\thetas}_i$. Under these conditions, and keeping all the terms that depend on the number of points in the approximation to the marginal likelihood, we have that
\begin{align}
    \dl(m_i, D) = \frac{B_1(m_i, D)}{2} - \log p(m_i) \;,
    \label{eq:dl_bic}
\end{align}
where $B_1(m_i, D)$ is the so-called Bayesian information criterion (BIC), and is given by \cite{schwarz78}
\begin{align}
    B_1(m_i, D) = - 2 \log p(D|m_i, \hat{\thetas}_i) + (k_i+1) \log N  \label{eq:bic}
\end{align}
with $k_i$ being the number of parameters in model $m_i$ (that is, the dimension of $\thetas_i$)\footnote{Note that the +1 in the term $(k_i+1)$ arises from the other parameter of the probabilistic model, that is, $\sigma$.} and
\begin{align}
    - \log p(D|m_i, \hat{\thetas}_i) =  \frac{N}{2} \left[ \log 2\pi + \log \frac{\sum_k \left( y^k - m_i(\xs^k, \hat{\thetas}_i) \right)^2}{N} +1 \right]
    \label{eq:loglike}
\end{align}
being the log-likelihood calculated at the maximum likelihood estimator of the parameters (including $\sigma$).

Adding an additional term to the approximation, we have 
\begin{align}
    \dl(m_i, D) = \frac{B_2(m_i, D)}{2} - \log p(m_i) \;,
    \label{eq:dl_gbic}
\end{align}
where $B_2(m_i, D)$ is given by
\begin{align}
    B_2(m_i, D) = - 2 \log p(D|m_i, \hat{\thetas}_i) + (k_i+1) \log N + \log \left| \mathcal{I}(\hat{\thetas_i}) \right| \,,
    \label{eq:gbic}
\end{align}
where $\mathcal{I}(\hat{\thetas_i})$ is the Fisher information matrix, calculated at the maximum likelihood estimators of the parameters \cite{ando10,bartlett2024}. The Fisher information matrix represents the curvature of the likelihood around its maximum at $\hat{\thetas}_i$, so that models with small curvature (that is, models for which changes in parameter values produce small changes in model predictions for observed data points) are preferred over models with a large curvature.

\section{Interpretations of the Bayesian approach}

\subsection{Probabilistic interpretation}

The probabilistic interpretation of the symbolic regression approach outlined above should be clear---each expression $m_i$ has a probability $p(m_i|D)$ of being the true generating model, and the most plausible model $\hat{m}$ is the maximum {\em a posteriori}
\begin{align}
    \hat{m} = {\arg \max}_{m_i} p(m_i|D) \,.
\end{align}
In the probabilistic interpretation, the posterior probability $p(m_i|D)$ is obtained by updating our prior expectations $p(m_i)$ about models with the marginal likelihood
\[
\int_{\Theta_i} {\rm d}\thetas_i \, p(D|m_i, \thetas_i) \, p(\thetas_i|m_i) \;,
\]
which can be approximated leading to Eqs.~\eqref{eq:dl_bic} and \eqref{eq:dl_gbic}.

Two important considerations, in this respect. First, the prior $p(m_i)$ does play a role in estimating the posterior $p(m_i|D)$---ignoring the prior amounts to assuming that all models $m_i$ are, in principle, equally plausible; and since there are exponentially many more complex models than simple models, it amounts to assuming that, in principle, complex models are more plausible than simple models. That being said, since the prior is fixed (intensive) and the marginal likelihood grows linearly with the number of observations $N$ (is extensive), in the limit $N\rightarrow\infty$ the prior washes out; that is, asymptotically, our prior expectations do not matter (as long as we do not assign $p(m_i)=0$ to any model).

Second, our prior expectations get updated by the integrated marginal likelihood, not the maximum likelihood or any other point estimate of the likelihood. This is important because a model $m_i$ may fit the data well for a specific choice $\hat{\thetas}_i$ of parameters, but poorly for other choices; and, since we are not certain about the exact values of the parameters, all values, good and bad, should be taken into consideration when evaluating the plausibility of the model. This is what happens, for example, to models with many parameters---one may find a good combination $\hat{\thetas}_i$, but the volume of models with poor fit grows with the dimension of the parameter space. This is the origin of the term that scales with the number of parameters $k_i+1$ in $B_1$ and $B_2$, which are sometimes interpreted as heuristic regularization terms but are, as we have seen, unavoidable consequences of the application of probability theory.

\subsection{Information-theoretic interpretation}

From Eq.~\eqref{eq:posterior}, one can see that 
\begin{align}
    \dl(m_i, D) = -\log p(m_i, D) = -\log p(D|m_i) - \log p(m_i) \;,
\end{align}
where $p(D|m_i)=\int_{\Theta_i} {\rm d}\thetas_i \, p(D|m_i, \thetas_i) \, p(\thetas_i|m_i)$ is the marginal likelihood.

In information-theoretic terms, $\dl(m_i, D)$ is the description length, that is, the number of nats (or bits, if we used base-2 logarithms instead of natural logarithms) necessary to convey $m_i$ and the data $D$ to a receiver by means of an optimal code \cite{grunwald07}. Then, from Eq.~\eqref{eq:posterior}, it is clear that the most plausible model $\hat{m} = {\arg \max}_{m_i} p(m_i|D) = {\arg \min}_{m_i} \dl(m_i,D)$ is the one with the minimum description length. This means that $\hat{m}$ is the model that allows the sender to convey that data most efficiently, that is, the model that best compresses the data.

The description length has two parts. The term $-\log p(m_i)$ corresponds to the number of nats necessary to convey model $m_i$ (among all possible models). The more plausible the model is {\em a priori}, the smaller this term. The second term $-\log p(D|m_i)$ corresponds to the number of nats necessary to convey the data once model $m_i$ is specified. If $m_i$ provides a better description of the data, then we need fewer nats in addition to the model itself to describe the data. Importantly, both terms are in the same ``units'' and are therefore comparable---description length nats provide a unified measure of model complexity and goodness of fit.

\subsection{Statistical physics interpretation}

Finally, the Bayesian approach can also be interpreted in the context of the canonical ensemble in statistical physics. Indeed, consider a physical system whose configurations are $c_i$. The probability of observing configuration $c_i$ in such a system is given by the Boltzmann distribution
\begin{align}
 p(c_i) = \frac{\exp \left[ -\beta{\mathcal H}(c_i) \right]}{Z}   
\end{align}
with $Z=\sum_j \exp \left[ -\beta{\mathcal H}(c_j) \right]$ being the partition function, ${\mathcal H(c_i)}$ being the energy of configuration $c_j$, and $\beta$ being the inverse of the temperature.

Then, by Eq.~\eqref{eq:posterior}, each model $m_i$ in symbolic regression can interpreted as a configuration whose energy is $\dl(m_i, D)$, for a system at temperature $\beta=1$. In the context of information field theories, $\dl$ is called the information Hamiltonian \cite{enslin2019}. In this picture, the most plausible model $\hat{m} = {\arg \max}_{m_i} p(m_i|D) = {\arg \min}_{m_i} \dl(m_i,D)$ is the one with the lowest energy, that is, the ground state of the system.

\section{Arguments for a Bayesian approach}

Consider a situation in which one draws a model $m \in \mathcal{M}$ from a distribution $p(m)$, and generates data according to Eq.~\eqref{eq:datagen}. Then, provided that the distribution $p(m)$ is known and used as the prior, the Bayesian approach in Eq.~\eqref{eq:posterior} is Bayes optimal, that is, it achieves the best possible expected performance and  no other algorithm can outperform it on average. Given the different axiomatizations of probability theory, this statement translates into different arguments for the use of the Bayesian approach as opposed to heuristic approaches.

\paragraph{Cox-type argument} Cox's theorem establishes that any system of reasoning under uncertainty that satisfies certain basic consistency and common sense requirements must necessarily follow the laws of probability theory \cite{cox46, jaynes03}. Therefore, it justifies the use of probability as the unique consistent framework for quantifying degrees of belief. Therefore, any way to assign plausibilities to models that does not conform to Eq.~\eqref{eq:posterior} must violate some of the very basic commonsensical conditions assumed by Cox.

\paragraph{Consistency argument} Related with the previous argument, the Bayesian approach outlined above is consistent, that is, in the large $N$ limit will select the true model with probability approaching one. In fact, this is true even if the prior $p(m)$ is unknown, because the marginal likelihood is extensive in $N$, whereas the prior is intensive. Therefore, any alternative that does not coincide with the Bayesian approach in this limit is virtually guaranteed to select the wrong model.

\paragraph{Minimum description length argument} As discussed above, the Bayesian approach selects the model with the shortest description length, that is, the model that maximally compresses the data. Any alternative way of selecting models will lead to models that compress the data less, that is, models that are objectively less parsimonious than those selected by the Bayesian approach.

\paragraph{Dutch book argument} In de Finetti's axiomatization, a probability is one's degree of belief in an event's occurrence, quantified as the price they would be willing to spend on a fair bet that pays one unit on the occurrence of the event \cite{dutch-book}. In this context, a Dutch book is a set of bets constructed to exploit non-probabilistic beliefs, guaranteeing a sure loss to those not using probability theory, no matter how the events unfold. Betting on symbolic regression models using any assignment of plausibility other than the Bayesian approach results in Dutch books, that is, in certain loss.

\section{Traditional heuristic approaches under the light of the Bayesian approach}

We hope that the reader finds the arguments in favor of the Bayesian approach in the previous sections convincing. However, one may still wonder how important these considerations are in practical terms. Here, we address this question by comparing the Bayesian approach to traditional heuristic symbolic regression approaches, both on theoretical grounds and in two simple scenarios.

We start by outlining how traditional symbolic regression works. First, an arbitrary loss function is defined, typically the squared error, which under the assumptions we have made here is equivalent to maximizing the likelihood $p(D|m_i, \hat{\thetas}_i)$ in Eq.~\eqref{eq:loglike}. Second, some algorithm, typically a genetic algorithm, is used to find models that minimize the loss. However, given a dataset $D$, the likelihood can always be made arbitrarily close to one by considering arbitrarily complex models. To escape this `structural overfitting,'\cite{guimera20} traditional symbolic regression proceeds by defining an heuristic measure of complexity, typically related to the number of operations and/or parameters in the model. Based on this complexity, one defines a Pareto front comprising the models that have minimum loss at each value of complexity. Finally, among all models in the Pareto front, one is typically selected by identifying (again, heuristically) the ``elbow'' of the front, that is, the point at which, somehow, the loss increases maximally for a fixed reduction in complexity.

All in all, the traditional approach involves three heuristic choices: loss, complexity and model selection criterion within the Pareto front (elbow). By contrast, the Bayesian approach does not require a heuristic definition of loss---the description length $\dl(m_i, D)$ is the quantity to minimize (or, equivalently, the posterior $p(m_i|D)$ is the quantity to maximize) as prescribed by probability theory. Similarly, there is no need to select models within the Pareto front because, as we have argued, $\dl(m_i, D)$ already combines goodness of fit and complexity within a single metric. The term quantifying goodness of fit is the marginal likelihood $-\log p(D|m_i)$.\footnote{The term $(k_i + 1) \log N$ in Eqs.~\eqref{eq:bic} and \eqref{eq:gbic} comes from approximating the marginal likelihood. Therefore, although often interpreted as a penalty to parametric complexity, it seems more appropriate to consider this term as part of the goodness of fit.} The term quantifying complexity is the prior $-\log p(m_i)$. Here, we follow previous work using a maximum entropy prior that reproduces the frequency of each mathematical operator $o\in \{+, \times, \exp, \log, \sin, \cos\dots\}$ in an empirical corpus of mathematical formulas \cite{guimera20}, as well as fluctuations of these frequencies, namely
\begin{align}
    p(m_i) = \exp \left[- \sum_{o} \left( \alpha_o n_{oi} + \beta_o n_{oi}^2 \right) \right] \,,
\end{align}
where $n_{oi}$ is the number of times that operator $o$ appears in $m_i$. In a sense, this choice of prior is arbitrary, but unlike in traditional approaches it is explicit and transparent, in the sense that all assumptions are explicit. Additionally, one could select other reasonable priors, including more informative priors encoding available background knowledge in a given context \cite{barlett23,fox2024}.

\begin{figure}[t!]
    \includegraphics[width=\linewidth]{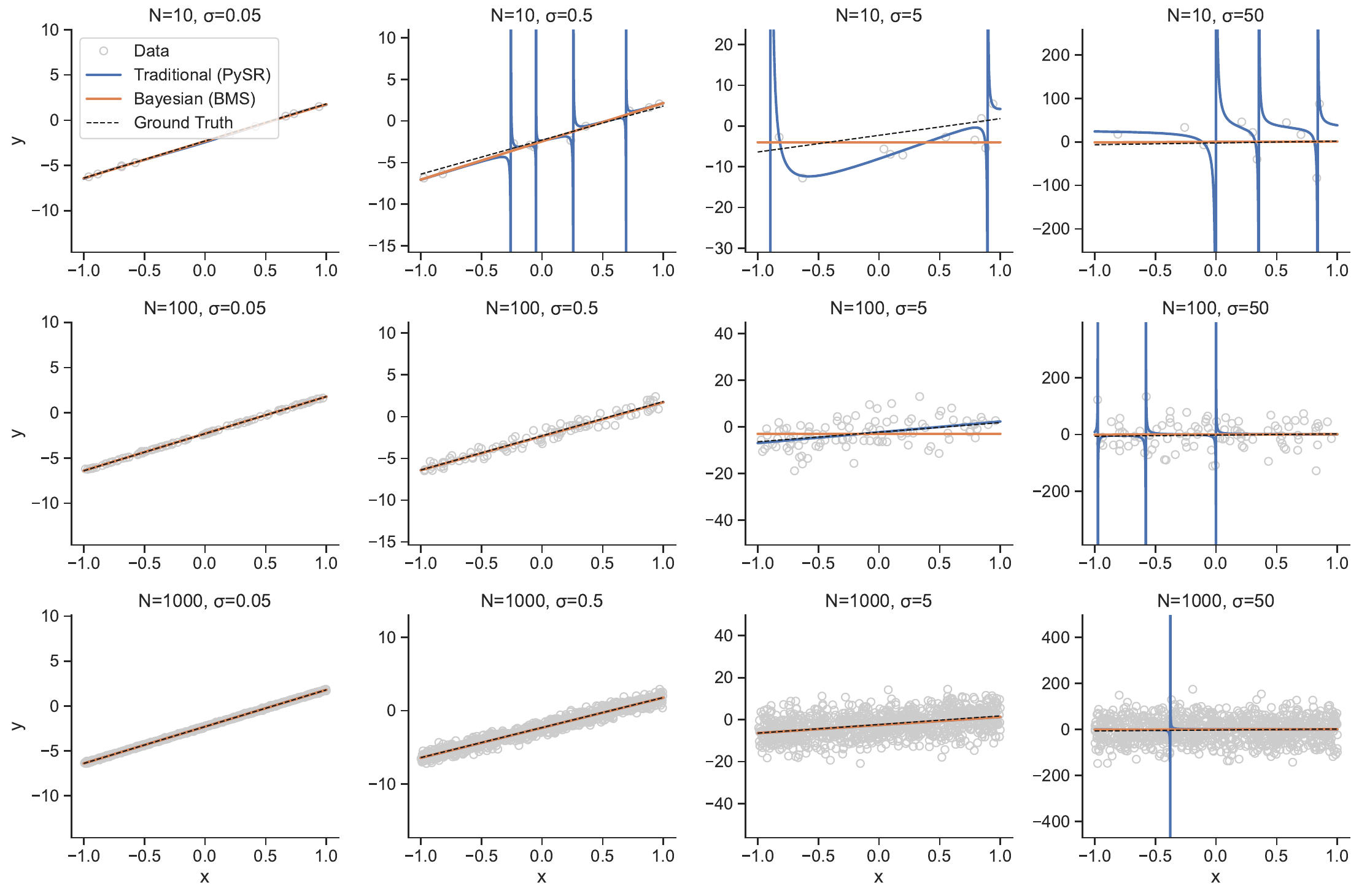}
    \caption{For varying number of data points $N \in \{10, 100, 1000\}$ and different levels of observational noise $\sigma \in \{0.05, 0.5, 5, 50\}$, we generate  data (gray symbols) through the process $y^k= \theta_0^* + \theta_1^* x + \epsilon^k$, with $\theta_0^*=-2.3$ and $\theta_1^*=4.1$. We then use traditional heuristic symbolic regression (using PySR \cite{pysr} with default parameters; blue lines) and Bayesian symbolic regression (using the Bayesian machine scientist, BMS \cite{guimera20}; orange lines) to learn $m^*$.}
    \label{fig:linear}
\end{figure}
To compare traditional and Bayesian symbolic regression in practice, we consider two simple scenarios. In the first one, we generate  data through the process $y^k= \theta_0^* + \theta_1^* x + \epsilon^k$, so that $m^*(\xs^k, \thetas^*) = \theta_0^*+\theta_1^* x$, with $\theta_0^*=-2.3$ and $\theta_1^*=4.1$. We then use traditional symbolic regression (using PySR \cite{pysr} with default parameters) and Bayesian symbolic regression (using the Bayesian machine scientist available at \url{https://bitbucket.org/rguimera/machine-scientist/}) to learn $m^*$. We repeat the process for varying number of data points $N \in \{10, 100, 1000\}$ and different levels of observational noise $\sigma \in \{0.05, 0.5, 5, 50\}$ (Fig.~\ref{fig:linear}).

This is a very simple model, where one may expect symbolic regression to work. Indeed, the Bayesian approach generally identifies the correct model---although in the high-noise regime it underfits the data, in those cases the error between the identified model and the true model (reducible error) is very small compared to the noise level $\sigma$ (irreducible error), so underfitting is actually reasonable. In practice, with such observational noise, making predictions with the true model or the underfit model would lead to almost identical errors, because error is dominated by the irreducible error $\sigma$.

The traditional approach also learns the correct model in the low-noise regime. However, when noise is high it overfits the data dramatically, even when the number of points is large; and, in this case, the reducible error is not necessarily small compared to the observational noise. The tendency of the traditional approach to overfit can be understood under the light of the Bayesian approach. Indeed, as we have argued, probability theory dictates that we select models by minimizing the description length
\begin{align}
    \dl(m_i, D) = - \log p(m_i) - \log p(D|m_i, \hat{\thetas}_i) + \frac{(k_i+1)}{2} \log N + \dots
\end{align}
where the dots indicate additional terms in the approximation of the exact marginal likelihood. In practice, traditional symbolic regression aims to minimize squared errors and, thus, to maximize the likelihood, so that the loss is
\begin{align}
    \dl_{\rm trad}(m_i, D) = \log p(D|m_i, \hat{\thetas}_i)\,.
\end{align}
By comparing the last two expressions, we note that the prior effectively being used by the traditional approach is 
\begin{align}
    \log p_{\rm trad} = \frac{(k_i+1)}{2} \log N + \dots \,,
\end{align}
that is, the traditional approach is favoring {\em a priori} models with {\em more}, rather than fewer, parameters. In fact, traditional approaches favor everything that the successive approximate terms of the marginal likelihood {\em penalize}. 

\begin{figure}[t!]
    \includegraphics[width=\linewidth]{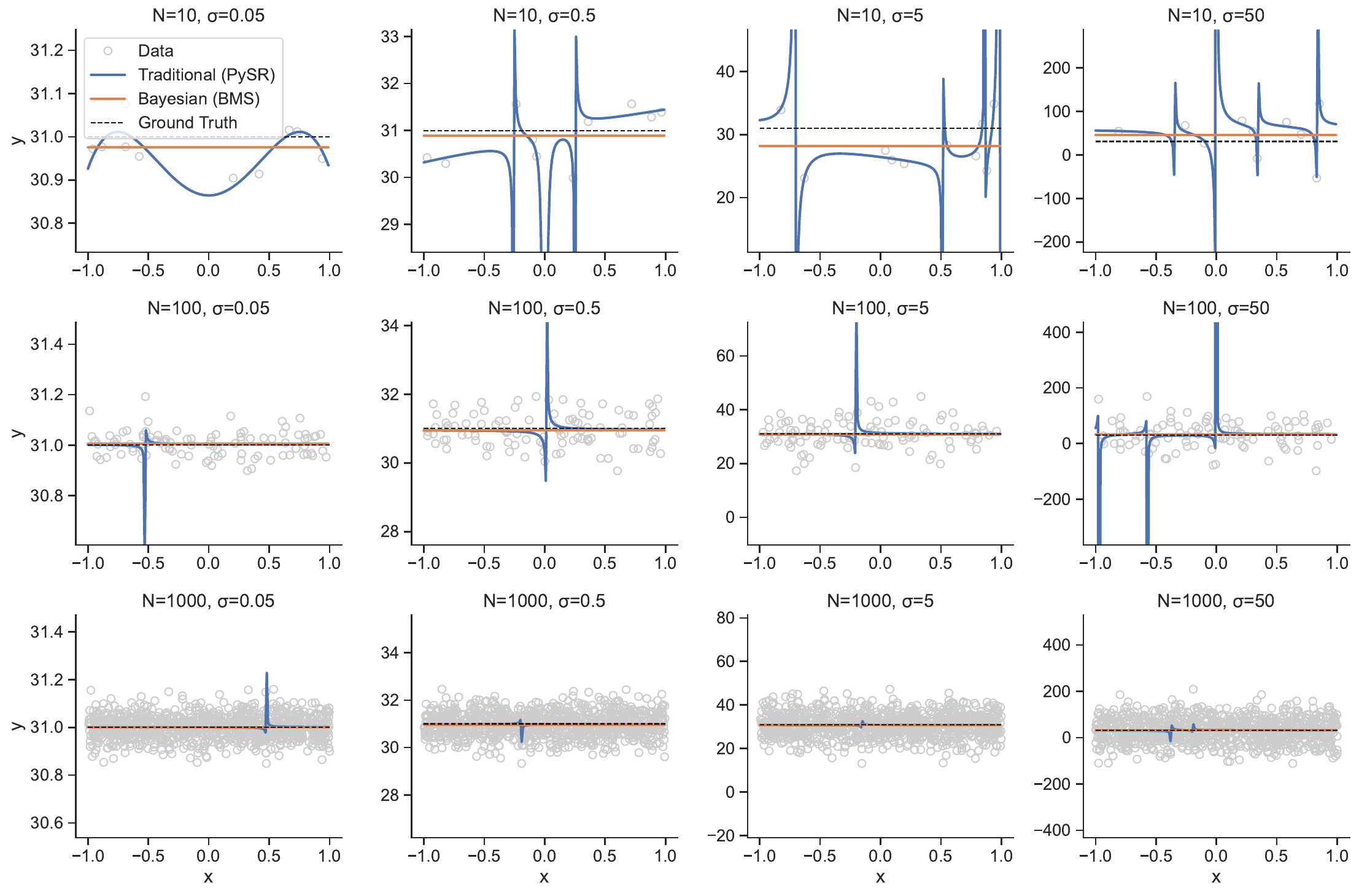}
    \caption{For varying number of data points $N \in \{10, 100, 1000\}$ and different levels of observational noise $\sigma \in \{0.05, 0.5, 5, 50\}$, we generate  data (gray symbols) through the process $y^k= \theta_0^* + \epsilon^k$, with $\theta_0^*=31$. We then use traditional heuristic symbolic regression (using PySR \cite{pysr} with default parameters; blue lines) and Bayesian symbolic regression (using the Bayesian machine scientist, BMS \cite{guimera20}; orange lines) to learn $m^*$.}
    \label{fig:constant}
\end{figure}
In the second experiment, we generate data through an even simpler process $y^k= \theta_0^* + \epsilon^k$, so that $m^*(\xs^k, \thetas^*) = \theta_0^*={\rm const.}$, with $\theta_0^*=31$ (Fig.~\ref{fig:constant}). Bayesian symbolic regression always identifies the correct model, although noise leads to estimates of the parameter $\hat{\theta}_0$ that deviate from the exact real value. By contrast, traditional symbolic regression fails to identify the correct model in every single instance and systematically overfits the data, even when noise is low and the number of points is high.

Besides the {\em a priori} preference for more complex models discussed in the previous experiment, the reason for the overfitting in this case is the heuristic used to select the best model within the Pareto front---since the constant model is the simplest possible model, it sits at the edge of the front and can never be considered an elbow. However, a linear model with a very small slope is also in the front and could, in principle, be selected. Rather, much more complex models are chosen in all cases, which means that, in this popular implementation of traditional symbolic regression, some complex model is always selected, even when no relationship whatsoever exists between dependent and independent variables. This example is thus sufficient to prove that the heuristics chosen do not lead to consistent model selection.

\section{From single models to posterior distributions over models}

So far, in line with traditional symbolic regression, we have focused on discussing how to get a single best model for a given dataset. In the Bayesian approach, we have identified this best model with the maximum {\em a posteriori}/minimum description length $\hat{m} = {\arg \max}_{m_i} p(m_i|D) = {\arg \min}_{m_i} \dl(m_i,D)$. However, the Bayesian approach does not only give the most plausible model, it gives the whole posterior $p(m_i|D)$, which contains much more information than any single model $m_i \in \mathcal{M}$.

\subsection{Model averaging}

Consider a situation in which one whats to predict the value of $y$ for a certain value of $\xs$, given the observed data $D$. One common approach to do this is to use the most plausible model and predict $y=\hat{m}(\xs, \hat{\thetas})$. However, it is important to note that this is just an approximation, because, in general, $p(\hat{m}|D) \ll 1$, that is, we have no certainty whatsoever that $\hat{m}(\xs, \hat{\thetas})$ is the true generating model. The statistical physics interpretation helps understand how incorrect this point estimate typically is---trying to predict $y$ with model $\hat{m}(\xs, \hat{\thetas})$ alone is like trying to predict the properties of a physical system using only the ground state configuration.

Rather, the most complete description of $y$ at $\xs$ is given by the posterior obtained through model averaging (or ensemble averaging) \cite{hoeting1999} 
\begin{align}
    p(y|D, \xs) = \sum_{m_i} \int_{\Theta_i} {\rm d}\thetas_i \, \delta\left(y - m_i(\xs, \thetas_i) \right) \, p(m_i, \thetas_i|D) \, ,
\end{align}
which is hard to calculate but can be approximated reasonably by
\begin{align}
    p(y|D, \xs) \approx  \sum_{m_i} \delta\left(y - m_i(\xs, \hat{\thetas_i}) \right) \, p(m_i|D) \approx \frac{1}{K} \sum_{m_i}{'} \delta\left(y - m_i(\xs, \hat{\thetas_i}) \right) \, .
\end{align}
Here as before, $\hat{\thetas}_i$ is the maximum likelihood estimator of the parameters of model $m_i$, and the primed sum $\sum{'}_{m_i}$ indicates that the sum is over $K$ models sampled from $p(m_i|D)$ using, for example, Markov chain Monte Carlo (MCMC) \cite{guimera20}.

\subsection{Fundamental limits and Rashomon sets}

Thinking about model averaging and model ensembles in the terms we have just discussed opens the door to deep and important questions about model space and the description length landscape. For example, is the true generating model always the ground state? And under what conditions is there a single model $\hat{m}$ that is overwhelmingly more plausible than any other model? Or, conversely, when do we have multiple models with description length similar to the ground state? 

Regarding the first question, analysis of the description length landscape leads to the conclusion that the ground truth generating model $m^*$ does not always coincide with the ground state $\hat{m}$ \cite{fajardo23}. Let us see why. As we have argued above, the probabilistic approach is consistent, that is, it identifies the true generating model with probability tending to one in the limit $N\rightarrow\infty$---in this limit, we do have  $\hat{m}=m^*$. However, for finite $N$, we can increase the observational noise $\sigma$ and, intuitively, it seems reasonable that, at some point, $m^*$ will become undetectable. This is indeed the case; and, in fact, this learnability transition (from a phase in which the true model can be learned to a phase in which it cannot) is properly described by considering only two minima in the description length landscape, namely, the ground truth model $m^*$ and the trivial model $m^0={\arg \max}_{m_i} p(m_i)$ that maximizes the prior over models \cite{fajardo23}. Since, as we have argued, the probabilistic approach is Bayes optimal, proving the existence of such learnable-to-unlearnable transition amounts to proving that, above a certain noise level, no algorithm can possibly identify the ground truth model from observational data alone. This is a result that only an approach that meets te fundamental requirements we have previously outlined, like the Bayesian approach, could have possibly obtained.

With regards to the other questions, it turns out that often there exist many models with description lengths similar to the ground state $\hat{m}$ for a given dataset. In other contexts, such collections of similarly plausible and explanatory models have been called Rashomon sets \cite{rudin19}\footnote{After the movie {\em Rashomon}, in which Akira Kurosawa  concatenates a series of contradictory accounts of the same crime.}. In the learnable phase, all models with description length similar to $\hat{m}$ are similar to the ground truth, so the Rashomon set does not add much to the single best model $\hat{m}$. However, close to the learnability transition, a Rashomon set of diverse models emerges, which provide non-congruent descriptions of the same data \cite{fajardo23}. In empirical datasets, this situation seems to be the norm rather than the exception \cite{guimera20,reichardt20,cabanas25}.

\section{Conclusion}

Luís A. N. Amaral has recently argued that research on ``artificial intelligence needs a scientific method-driven reset'' based on reliable use of ``prior knowledge, falsifiable hypotheses, and rigorous experimentation'' \cite{amaral24}. This, we believe, is true in general but especially for applications of AI in science and for symbolic regression in particular.  

Here, we have compared probabilistic to traditional symbolic regression. Of course, in recent years there has been an explosion of new symbolic regression methods based on large language models, variational autoencoders, and a variety of other deep learning approaches. However, the main limitations we have identified and discussed here for traditional symbolic regression remain in these newer approaches, namely: (i) the need to define goodness of fit (or loss) and complexity measures heuristically; (ii) the need to choose models heuristically based on fit and complexity; and (iii) the need to explore model space heuristically. The probabilistic approach provides concrete and easy-to-implement alternatives to (i) and (ii), so we see no reason why all other approaches should not adopt them. With regards to (iii), heuristic search is acceptable for practical applications, but one must always keep in mind that, for certain advanced applications (such as model averaging for prediction, or analysis of model space for theoretical results like those related to learnability), sampling over the posterior distribution provides the most detailed description of the problem.

Symbolic regression can revolutionize the scientific process by automating the learning of closed-form mathematical models from data. However, for symbolic regression to advance on solid grounds, and to help other fields also advance on solid grounds by identifying models that are defensible, it must aim for the maximum levels of mathematical and conceptual rigor. In this manuscript, we have argued that the interface between probability theory, information theory and statistical physics provides the ideal framework for this.

\ack{This research was funded by project PID2022-142600NB-I00 from MCIN/ AEI/10.13039/501100011033 FEDER, UE, and by the Government of Catalonia (2021SGR-633).}


\bibliographystyle{RS}
\bibliography{ref-database,add_refs}

\end{document}